\documentclass[10pt,twocolumn,letterpaper]{article}

\usepackage{cvpr}              
\usepackage{url}

\definecolor{cvprblue}{rgb}{0.21,0.49,0.74}
\usepackage[pagebackref,breaklinks,colorlinks,allcolors=cvprblue]{hyperref}
\usepackage{multirow}
\def\paperID{*****} 
\def\confName{CVPR}
\def\confYear{2025}

\title{InstructVTON: Optimal Auto-Masking and Natural-Language-Guided Interactive Style Control for Inpainting-Based Virtual Try-On}

\author{
Julien Han\textsuperscript{1} 
\quad
Shuwen Qiu\textsuperscript{2}\thanks{Work done during internship at Amazon.}
\quad
Qi Li\textsuperscript{1}\quad
Xingzi Xu\textsuperscript{1,3 *}\quad
Mehmet Saygin Seyfioglu\textsuperscript{1}\quad \\
Kavosh Asadi\textsuperscript{1}\quad
Karim Bouyarmane\textsuperscript{1} \vspace{8pt}\\
\textsuperscript{1}Amazon \quad
\textsuperscript{2}University of California, Los Angeles (UCLA) \quad
\textsuperscript{3}Duke University
\vspace{8pt}\\
{\tt\small \{hameng,qlimz,xingzixu,mseyfiog,kavasadi,bouykari\}@amazon.com}\\
{\tt\small xingzi.xu@duke.edu} \\
{\tt\small jantqiu@cs.ucla.edu} \\
{\color{red}{\tt\small \url{https://instructvton.github.io/instruct-vton.github.io/}}} \\
{\small \textbf{Work submitted in November 2024 to CVPR 2025}}
}

\begin{document}
\maketitle
\begin{abstract}
We present InstructVTON, an instruction-following interactive virtual try-on system that allows fine-grained and complex styling control of the resulting generation, guided by natural language, on single or multiple garments. A computationally efficient and scalable formulation of virtual try-on formulates the problem as an image-guided or image-conditioned inpainting task. These inpainting-based virtual try-on models commonly use a binary mask to control the generation layout. Producing a mask that yields desirable result is difficult, requires background knowledge, might be model dependent, and in some cases impossible with the masking-based approach (e.g. trying on a long-sleeve shirt with “sleeves rolled up” styling on a person wearing long-sleeve shirt with sleeves down, where the mask will necessarily cover the entire sleeve). InstructVTON leverages Vision Language Models (VLMs) and image segmentation models for automated binary mask generation. These masks are generated based on user-provided images and free-text style instructions. InstructVTON simplifies the end-user experience by removing the necessity of a precisely drawn mask, and by automating execution of multiple rounds of image generation for try-on scenarios that cannot be achieved with masking-based virtual try-on models alone. We show that InstructVTON is interoperable with existing virtual try-on models to achieve state-of-the-art results with styling control. 
\end{abstract}
\section{Introduction}

\begin{figure*}[hbt!]
    \centering
    \includegraphics[width=\textwidth]{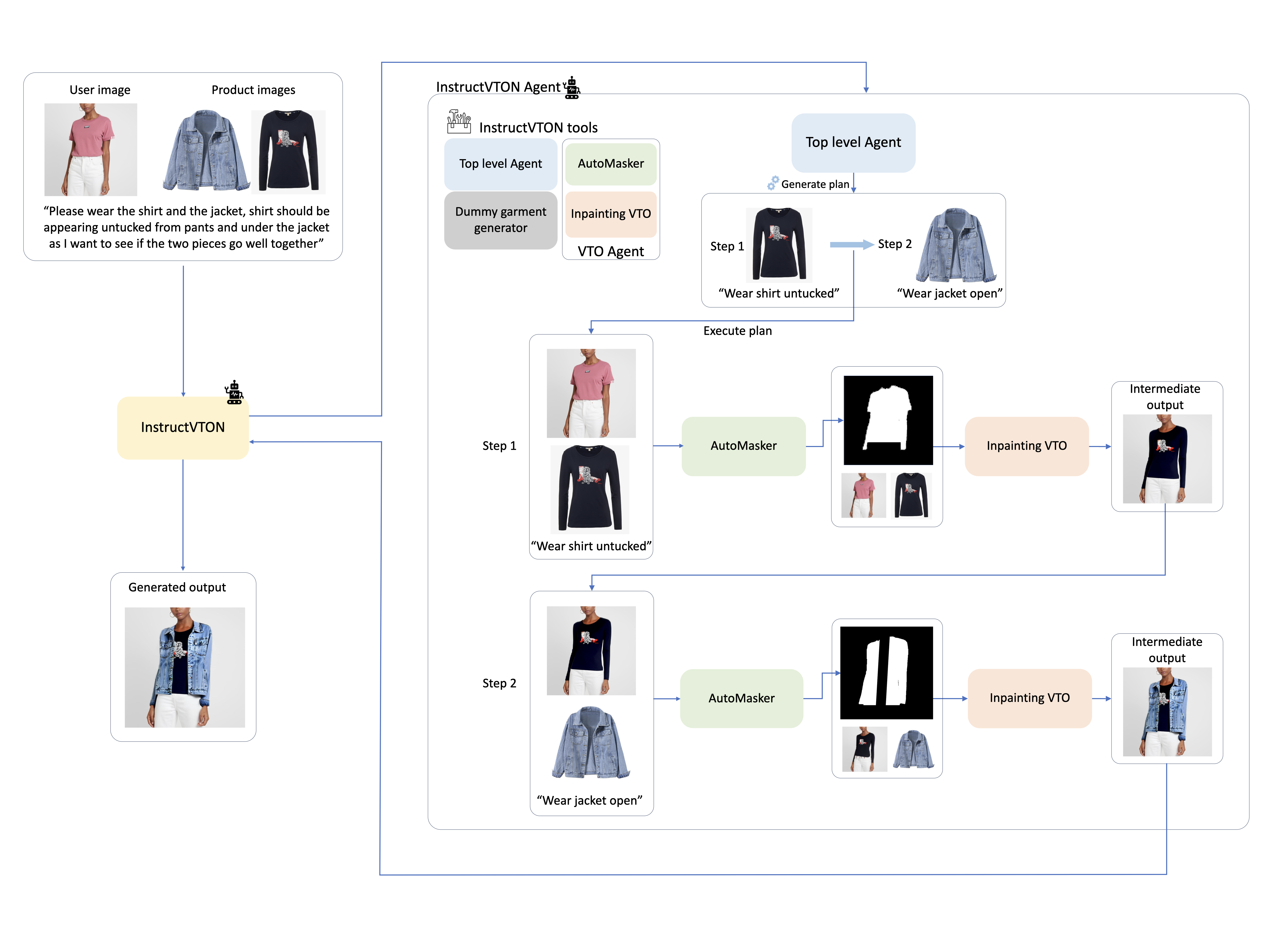}
    \caption{Overview of the InstructVTON architecture, top-level agent plans order to try-on each garment and summarizes corresponding style instruction, VTO agent performs masking and execute VTO model to obtain try-on results.}
    \label{fig:ins-vton}
\end{figure*}

Virtual try-on (VTO) is a use-case of image inpainting where the VTO model generates a photo-realistic image of a person wearing a target garment by combining an image of the person and an image of the target garment \cite{qi2024ditvton,xu2025deft,qi2024posevton,choi2021vitonhdhighresolutionvirtualtryon,chong2024catvtonconcatenationneedvirtual,han2018vitonimagebasedvirtualtryon,kim2023stablevitonlearningsemanticcorrespondence,xie2023gpvtongeneralpurposevirtual}. As a possible application, online shopping retailers can use the VTO models to allow customers to interactively try-on products before they make a purchase. VTO can be also seen as an image editing tool, where users can make use of the VTO models to generate images of human models wearing different products to create marketing content without needing to create images in a photo studio.

Recent advancements in Diffusion models \cite{dhariwal2021diffusionmodelsbeatgans,ho2020denoisingdiffusionprobabilisticmodels,song2022denoisingdiffusionimplicitmodels} are offering promising alternatives to traditional GAN-based approaches for virtual try-on (VTO) systems. Diffusion models have shown exceptional capabilities in generating high-quality and coherent images, which makes them particularly suitable for virtual try-on applications. Recent research in diffusion-based VTO focus on creating natural-looking try-on results, preserving the fine-grained details of the target garment, enabling styling control, supporting multi-garment try-on, and producing high-resolution images. 

One common and efficient formulation of the VTO problem in the literature is that of image-guided or image-conditioned inpainting task~\cite{qi2024ditvton,xu2025deft,qi2024posevton}. Inpainting-based VTO models share common requirements for their input, which typically comprise an image of the human model, an image of the target garment, and a binary mask to indicate where the garment will be positioned on the human model. However, creating this mask for end-users often requires an understanding of the working mechanism of the VTO model mechanics. Moreover, in some cases it is impossible to achieve the desired result with specific styling control with a single round of execution of the VTO model, for example, generating a try-on result of a long-sleeve shirt with sleeves rolled up on a human model wearing a long-sleeve shirt. Masking the long-sleeve shirt on the human model and provide the long-sleeve shirt as target garment will prompt the VTO model to generate a try-on image with target garment with sleeves rolled down, as there is no information about the desired sleeve-rolled up style in the input combination. These limitations of the current VTO models pose challenges to delivering a seamless and intuitive user experience.

To address these limitations, we propose InstructVTON, an automasking and agentic system which allows users to guide the inpainting process using natural language instead of a mask, providing a more intuitive experience interacting with the VTO system. InstructVTON also allows user to provide multiple garments at a time to enable an outfit try-on experience. Consider a complex case where a user would like to try on a shirt, a pair of pants and a jacket with the shirt tucked in the jacket buttons open. The user only needs to provide the image of the target garments, a human model image and an instruction \emph{“try on the shirt tucked in, jacket open”}, InstructVTON will automatically arrange the target garments in the correct try-on order and iteratively generate try-on results with each target garment to produce the final result. InstructVTON harnesses the visual reasoning capability of Vision Language Models (VLMs) and the power of image segmentation models to determine the series of actions to interact with inpainting VTO models and generate the appropriate mask based on the style instruction. Specifically, InstructVTON organizes multi-garment try-on cases into single-garment try-on by ordering the target garments in the correct order based on the style instruction, generates a structured action plan with the ordered target garments along with the corresponding style instruction summarized from the original style instruction from the user. For each target garment, InstructVTON determines an action plan to execute the VTO model given the style instruction, i.e. where in the human model image to mask, and whether multiple rounds of image generation are necessary to achieve the desired style. Each time InstructVTON executes the VTO model, it uses the segmentation maps of the human model image, the target garment image and style instruction as a basis to automatically determine the mask input that can achieve the try-on result following the style instruction, that we call AutoMasker. 

We summarize our contributions below:
\begin{itemize}[leftmargin=0.3in]
  \item We introduce a novel agentic system that autonomously executes a VTO model to achieve multi-garment virtual try-on with style control following free-text instruction provided by the user (InstructVTON). 
  \item We formally define and decompose the mask generation mechanism and integrate it into the agentic system to automatically determine the appropriate masking area given style instruction (AutoMasker).
  \item We conduct extensive experiments with state of the art VLMs, segmentation models and VTO models to demonstrate that InstructVTON is interoperable with existing VTO models without the need for retraining or fine-tuning.
\end{itemize}
\section{Related work}

\noindent\textbf{Virtual try-on.} Virtual try-on refers to the task that creates the image where a person naturally wears the provided garment
~\cite{choi2021viton,ge2021parser,lee2022high,xie2023gp}. With the advent of more powerful latent diffusion models~\cite{esser2024scaling, nichol2021glide, rombach2022high, saharia2022photorealistic}, attentions have shifted to how to effectively integrate the garment details into the person’s context, including dual UNet~\cite{zhu2023tryondiffusion}, encouraging sparse attention over the clothing area~\cite{kim2024stableviton}, texture and human identity preservation~\cite{choi2024improving, yang2024texture} and efficient training and inference~\cite{chong2024catvton}. Recently, VTO task was extended to multi-garment try-on~\cite{li2024anyfit}, style controlled try-on~\cite{zhu2024m}, and mask free try-on~\cite{zhang2024boow}. 

\noindent\textbf{Image segmentation.} Image segmentation was long formulated as a pixel classification task \cite{chen2018encoderdecoderatrousseparableconvolution,long2015fullyconvolutionalnetworkssemantic,zhao2017pyramidsceneparsingnetwork}, more recent work has adopted transformer models for instance-level segmentation \cite{cheng2022maskedattentionmasktransformeruniversal,cheng2021perpixelclassificationneedsemantic,jain2022oneformertransformerruleuniversal,liu2023simpleclickinteractiveimagesegmentation,zhang2019pose2segdetectionfreehuman}. Work of Kirillov et al. \cite{kirillov2023segment} introduced a segmentation model that can be used to segment any object, while human parsing models \cite{güler2018denseposedensehumanpose} \cite{khirodkar2024sapiensfoundationhumanvision} are developed to specialize in human related segmentation tasks such as pose estimation and body parts segmentation. 

\noindent\textbf{Vision language models and agents.} Vision language Models (VLMs) \cite{anthropic2024claude3,chen2024internvlscalingvisionfoundation,deitke2024molmopixmoopenweights,laurençon2024mattersbuildingvisionlanguagemodels,touvron2023llamaopenefficientfoundation} fuse visual understanding capabilities into language models and created the possibilities to create agents for visual tasks \cite{sumers2024cognitivearchitectureslanguageagents,xi2023risepotentiallargelanguage,xie2024largemultimodalagentssurvey}. Recent research has been done to build generalist autonomous agent \cite{zhang2024cognitivekernelopensourceagent} and agent that can interact with other systems via APIs \cite{liu2024summaryactionenhancinglarge}. Reinforcement learning has also been applied to improve decision making in agents \cite{golchha2024languageguidedexplorationrl}. 

\section{Method}

Figure \ref{fig:ins-vton} shows an overview of the InstructVTON architecture. InstructVTON comprises a Top level Agent and a VTO Agent. Given a human model image, a set of target garment images and a style control instruction, the Top level Agent organizes the multi-garment try-on task into a sequence of single-garment try-on tasks. It generates an action plan with the correct try-on order for the target garments and paraphrase the style control instruction for each target garment. The VTO Agent then executes the action plan step by step, the output from a step is used as the human model image input for the following step. At each step, the VTO Agent receives a human model image, a target garment image and a style instruction corresponds to the target garment as input. It uses the segmentation maps of the human model image from the AutoMasker as the basis to determine the masking area in the human model image that can satisfy the style instruction. Finally, it invokes the VTO model to generate the try-on image. The Top level Agent, VTO Agent, AutoMasker and inpainting VTO model interact using structured input and output. We discuss the details of the AutoMasker mask generation process, the Top level Agent and the VTO Agent below.

\subsection{AutoMasker: Optimal Auto-Mask Generation for instruction-guided inpainting VTO}

\begin{figure}
    \centering
    \includegraphics[width=\linewidth]{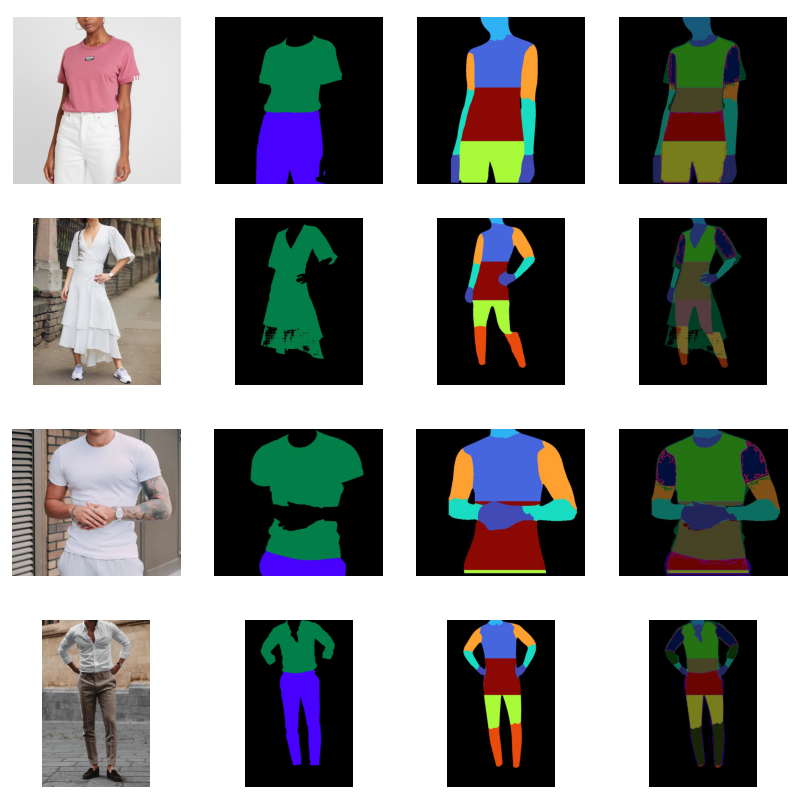}
    \caption{In each row from left to right are human model image, clothing segmentation map ($\mathcal{C}$), body parts segmentation maps ($\mathcal{B}$), and all pairwise intersections of $\mathcal{B}\times\mathcal{C}$, which shows the maximum approximation granularity our our AutoMasker. 
    }
    \label{fig:seg-map-b-c}
\end{figure}

\begin{figure*}[t]
    \centering
    \includegraphics[width=0.9\textwidth]{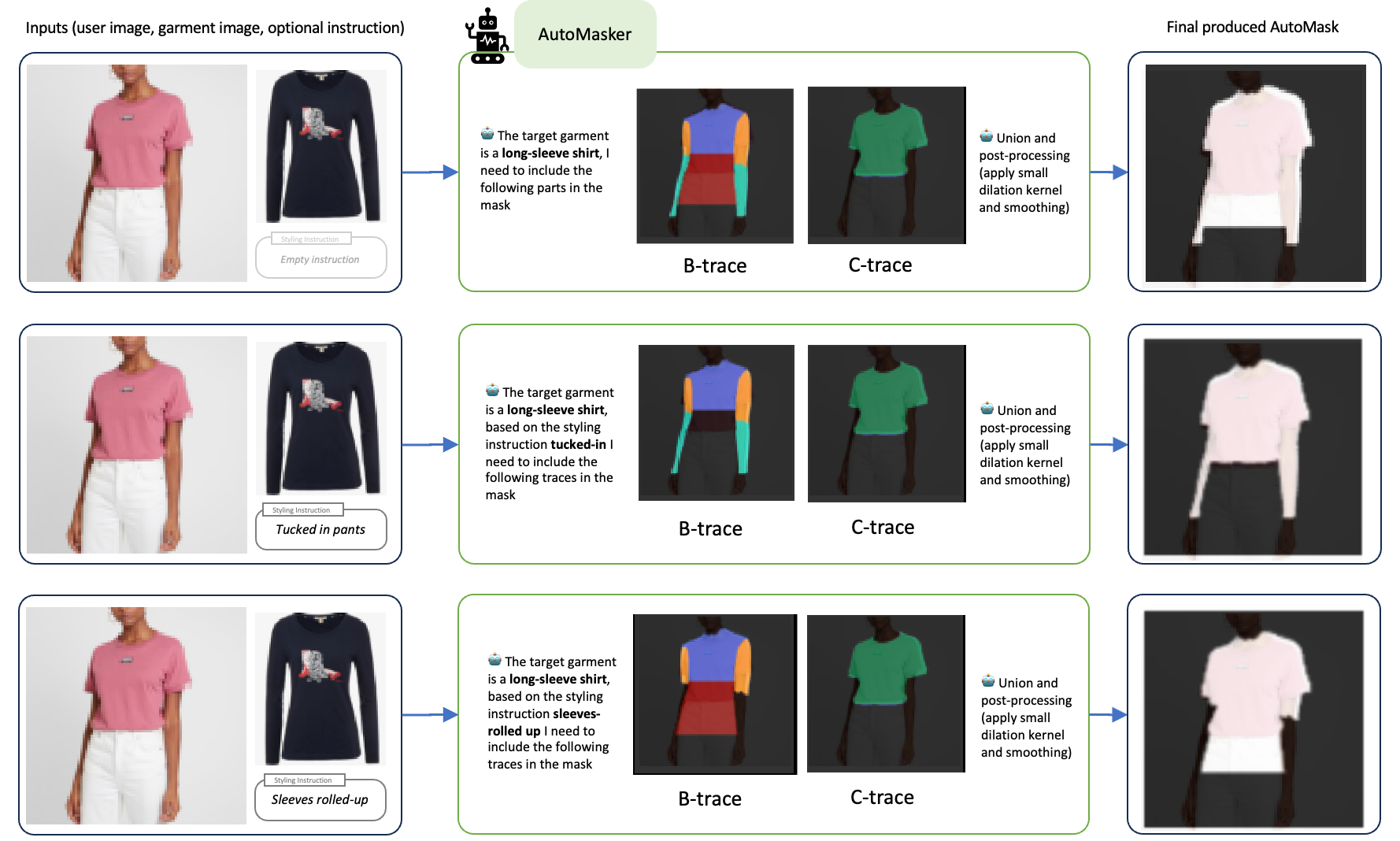}
    \caption{AutoMasker reasoning and masks produced by different instructions.}
    \label{fig:masks-instructions}
\end{figure*}

In inpainting-based VTO applications, the mask plays a crucial role in defining the inpainting area while preserving the surrounding content. The ideal mask should cover the smallest portion of the human model image that is necessary to achieve the desired effects. This will minimize the destruction of the source image. Typical auto-masking solutions for state-of-the-art VTO models produce masks based on target garment's top/bottom/overall classification. These masks typically overflow well beyond the area strictly necessary to achieve the desired try-on results, to be conservative and ensure that the existing garment on the person is completely removed. We propose an optimal, minimally-invasive auto-masking approach which minimizes the mask-to-image ratio (a concept we name "mask efficiency") while achieving the same try-on effect.

To achieve optimal mask efficiency, we use segmentation models that produce a segmentation map given an image. A body parts segmentation map (BPSM) model provides a segmentation map of human body parts in an image. A clothing segmentation map (CSM) model provides a segmentation map of the existing clothing in an image. 
Figure \ref{fig:seg-map-b-c} shows examples of segmentation maps from BPSM and CSM for the human model image. In our approach, we denote $\mathcal{B}=\{b_1,b_2,b_3,...\}$ the body part segmentation produced by BPSM, and $\mathcal{C}=\{c_0,c_1,c_2,c_3,...\}$ the clothing segmentation produced by CSM, where $c_0$ denotes the unclothed area. We denote $B$ the entire human figure and existing clothing in the human model image. Both $\mathcal{B}$ and $\mathcal{C}$ constitute partitions of $B$ (i.e. $\bigcup b_i = B, \forall i,j\ b_i\cap b_j=\emptyset, \bigcup c_i = B, \forall i,j\ c_i\cap c_j=\emptyset$). We denote $\bar{v}$ the ideal position of the target garment in the output image and we denote the following “traces”, B-trace (body parts segment trace): $\mathcal{B}_{\cap \bar{v}}=\{b_i\ |\ b_i\cap\bar{v}\neq\emptyset\}$ and C-trace (existing clothing segment trace): $\mathcal{C}_{\cap \bar{v}}=\{c_i\ |\ c_i\cap\bar{v}\neq\emptyset\}$. We denote $\bar{m}$ the optimal (minimally-invasive) masking area. We can observe that, $\bar{m}=\left(\bigcup_{c_i\in\mathcal{C}_{\cap \bar{v}}} c_i\right)\cup \bar{v}$, that is the union of the existing garment that need to be removed and the ideal position of the target garment. We do not know a priori $\bar{v}$, but we can use the type of the target garment and styling instruction to estimate $\mathcal{B}_{\cap \bar{v}}$ and $\mathcal{C}_{\cap \bar{v}}$, we denote the estimated traces as $\hat{\mathcal{B}}_{\cap \bar{v}}$ and $\hat{\mathcal{C}}_{\cap \bar{v}}$. Then we estimate the minimum invasive masking $\hat{m}= \left(\bigcup_{c_i\in\hat{\mathcal{C}}_{\cap \bar{v}}} c_i\right)\cup \left(\bigcup_{b_i\in\hat{\mathcal{B}}_{\cap \bar{v}}} b_i\right)$.

\begin{figure*}[h!]
    \centering
    \includegraphics[width=0.9\textwidth]{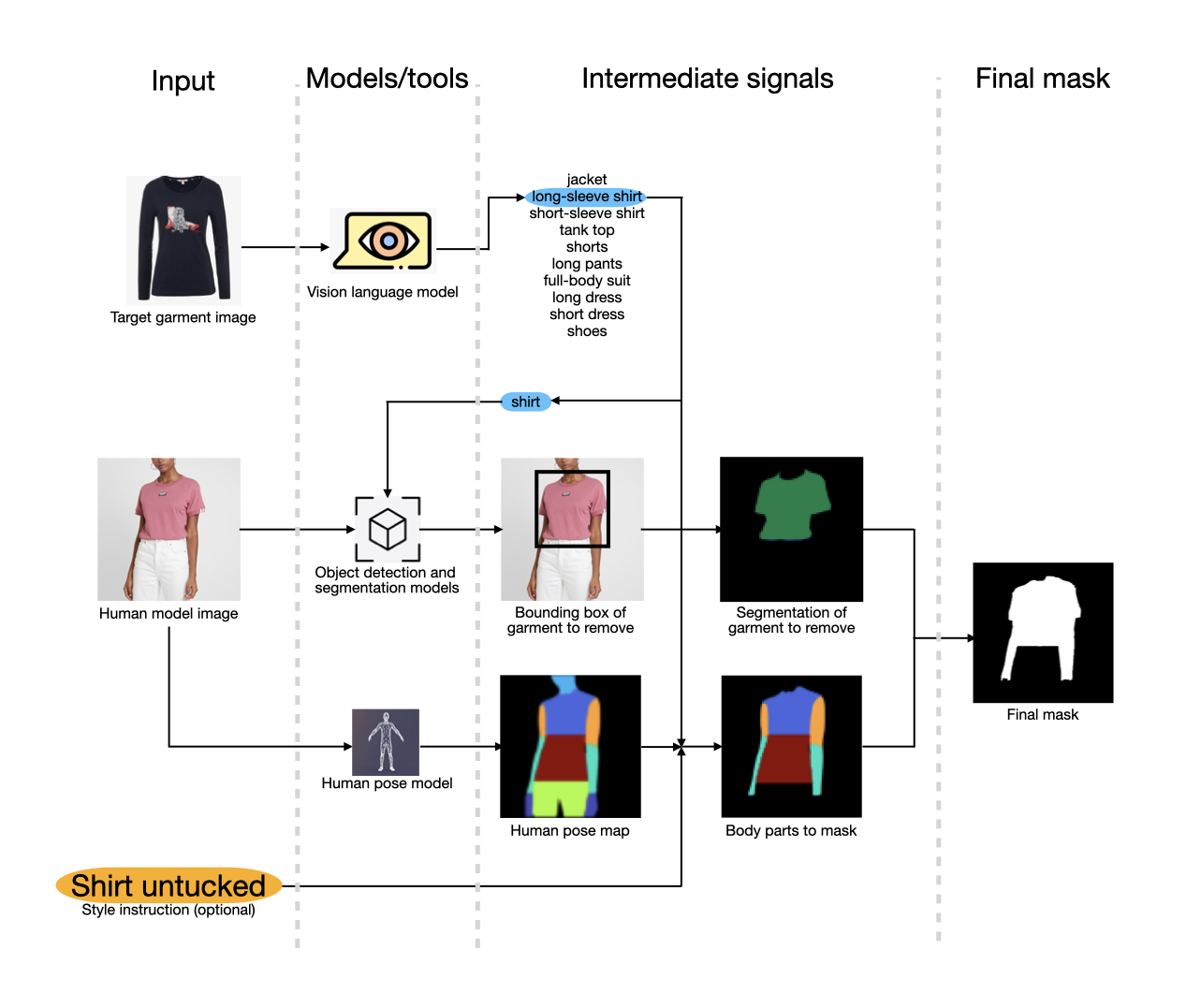}
    \caption{AutoMasker detailed architecture and agent tools.}
    \label{fig:automasker architecture}
\end{figure*}

Practically, BPSM produces a body part segmentation map that comprises face, upper torso, lower torso, upper arms, lower arms, hands, upper legs, lower legs and feet. CSM provides a segmentation of all clothing pieces. We define a set of \emph{parts inclusion rules} based on target garment classification (upper, lower, overall), target garment structured attributes (sleeve length, leg length, closure type, etc), and structured styling instructions. These rules allow us to decide which components from $\mathcal{B}$ and $\mathcal{C}$ to include in our estimated mask area $\hat{m}$, and if additional processing such as making the legs area convex when the target garment is a dress or a coat, or removing a stripe in the center of the torso if style instruction mentions an open-chest style. Figure \ref{fig:masks-instructions} shows different B-trace and C-trace generated by the parts inclusion rules given different styling instructions. The final mask is the union of B-trace and C-trace. For simplicity, we assume corresponding left and right limbs need to be masked simultaneously, although they can be masked separately to try-on asymmetrical garments. Figure \ref{fig:automasker architecture} shows the detailed architecture of AutoMasker. We compare the masking result of our approach and existing masking approaches in experiment section.


\subsection{InstructVTON Agent}

\begin{figure*}[ht!]
    \centering
    \includegraphics[width=0.9\textwidth]{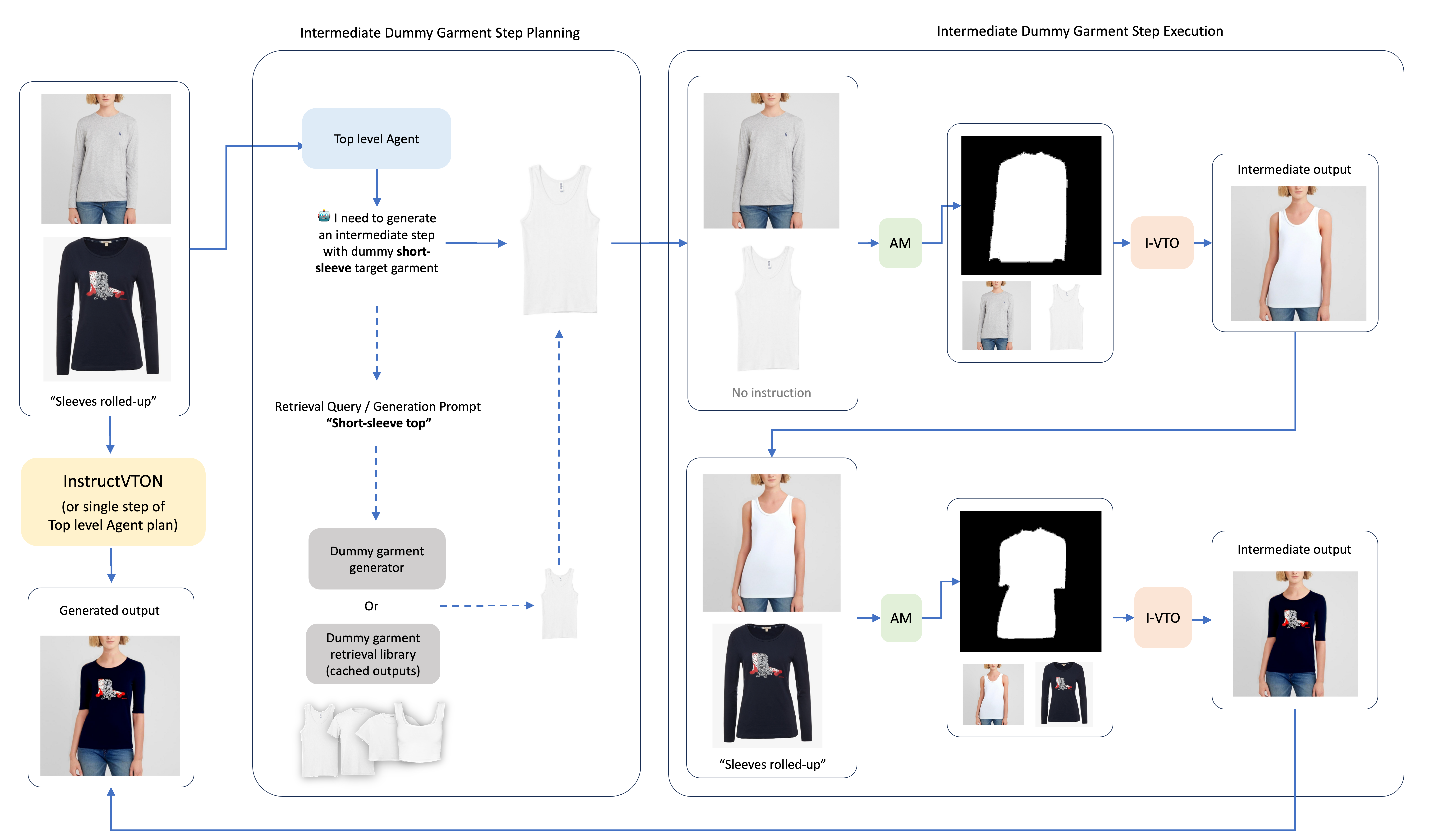}
    \caption{VTO Agent adopts a two-step approach with a generic tank top (dummy garment retireved from predefined library or generated from text-to-image generator) as intermediate target garment to first shorten the sleeves in the human model image, then uses a second step to generate try-on image satisfying style instruction.}
    \label{fig:tank-top-as-intermediate}
\end{figure*}

The VTO task involves understanding the human model image, the target garment image, and deciding the area to mask in the human model image to achieve the desired results. InstructVTON also needs to understand the style instruction and adjust its decisions about the area to mask. Certain styling instructions cannot be achieved with a single execution of the inpainting VTO model. For example, a human model image wearing a long-sleeve shirt and the style instruction is to try-on another long sleeve shirt with sleeves rolled-up, in this case the agent needs to execute the inpainting VTO model once with an alternative target garment image (e.g. a tank top) to generate an intermediate human model image with arms not covered by clothing. It is then possible to achieve the sleeves rolled up style with a second execution of the inpainting VTO model with an intermediate human model image and the original target garment. When there are multiple reference images, InstructVTON needs to decide on the correct order to generate the try-on result with each reference image. For example, try on a shirt before a jacket in open-chest style to layer the jacket the jacket on top of the shirt.

InstructVTON accepts a human model image $I_\text{src}$, a set of target garment images $S_\text{ref}$. Optionally, user can provide a free-text instruction $T_\text{instruction}$ to specify desired style for the try-on result. InstructVTON then plans and executes actions to generate final try-on result $I_\text{try-on}$. Practically, InstructVTON comprises 2 major components - Top level Agent and VTO Agent. The Top level Agent generates an execution plan by reordering the target garment images in $S_\text{ref}$ and summarizing the style instruction for each target garment from $T_\text{instruction}$. The VTO Agent executes the plan step by step. When executing the first step, it uses the original $I_\text{src}$ as human model image, at each proceeding step, it uses the output from the previous step as human model image. Output from the last step is the final try-on result $I_\text{try-on}$. 


In certain cases, the style instruction cannot be satisfied by directly inpainting the target garment into the human model image. For example, try on a long-sleeve shirt target garment with a human model image wearing a long-sleeve shirt following style instruction “sleeve rolled up”. The masking area needs to cover the entire upper clothing in the human model image, as the existing clothing needs to be removed. Although the inpainting VTO model will not generate a try-on image of a long-sleeve shirt with sleeves rolled up given this mask. A solution is to use an alternative target garment with short sleeves to lead the inpainting VTO model to generate an intermediate human model image with arms exposed, which is then paired with the original target garment to generate the final try-on result that satisfies the style instruction with a mask that does not cover the lower arms. The alternative garment (we call "dummy garment") is fetched from a library or synthesized from a "Dummy-garment generator". Figure \ref{fig:tank-top-as-intermediate} shows an example of InstructVTON using a dummy tank top image as an alternative target garment to generate an intermediate try-on result, then uses the intermediate try-on result as human model image to product final try-on result that satisfies the style instruction.

\section{Experiments}

\subsection{Mask efficiency and try-on result quality comparison}

We provide in this section details on mask efficiency metric (mask optimality) and quality of try-on results of InstructVTON. Mask efficiency is measured by the ratio of the human model image covered by the mask.
$$
\text{Mask efficiency} = 1-\frac{\text{masked area}}{\text{total image area}}
$$

An optimal auto-masker should be able to achieve higher mask efficiency by preserving as many unmasked pixels as possible, and should try to keep the mask to the strictly necessary area to achieve the desired effect. Given comparable try-on results, it is more desirable to mask smaller ratio of the human model image as more pixels from the original image will be preserved. A trivial auto-masker could, for example always mask the entire image while keeping the face of the subject.

\begin{figure}
    \centering
    \includegraphics[width=0.9\columnwidth]{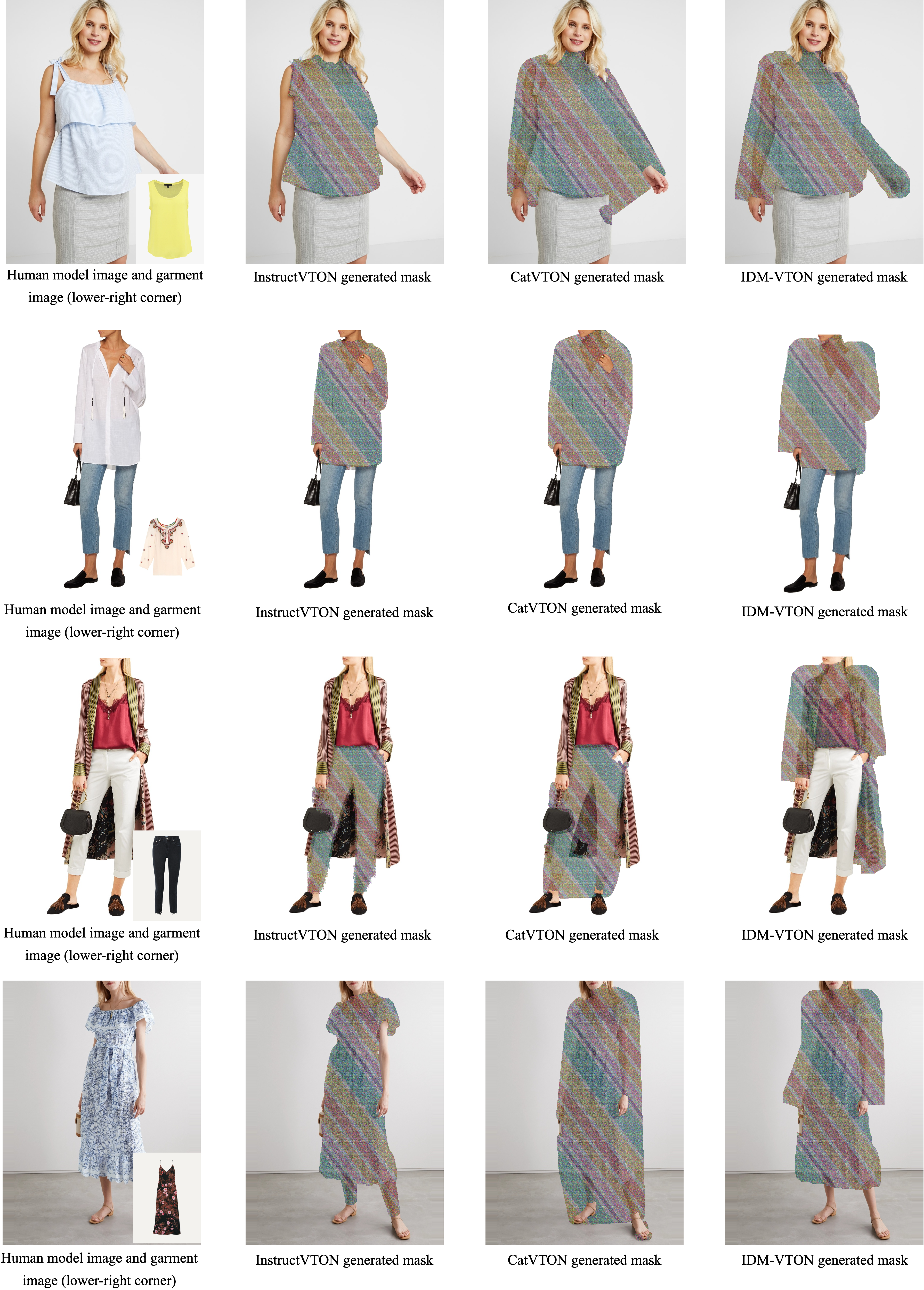}
    \caption{Examples of masks generated by InstructVTON, CatVTON and IDM-VTON. InstructVTON generates masks that only covers the strictly necessary area.}
    \label{fig:mask-examples}
\end{figure}

To quantitatively measure the quality of the try-on result that constrains the mask efficiency, we use Structural Similariy Index Measure (SSIM) and Learned Perceptual Image Patch Similarity (LPIPS). We randomly sample 50 pairs of human model image and clothing image from each of the 3 categories (dresses, upper body, lower body) in the public DressCode dataset, and 100 pairs from the public VITON-HD dataset. For all the examples, InstructVTON is instructed with empty instruction. Table \ref{tab:mask-efficiency} shows the mask efficiency compared to two state-of-the-art open-source VTO model baselines that provide an auto-masking feature (IDM-VTON\cite{choi2024improving} and CatVTON~\cite{chong2024catvtonconcatenationneedvirtual}). Figure \ref{fig:mask-examples} shows examples of masks generated by InstructVTON, CatVTON and IDM-VTON. InstructVTON consistently achieves higher masking efficiency. Table \ref{tab:ssim} and Table \ref{tab:lpips} show that InstructVTON maintains or improves image generation quality when compared to CatVTON and IDM-VTON.

\begin{table}[h!]
\centering
\resizebox{\linewidth}{!}{
\begin{tabular}{ccccc}
\toprule
\multirow{2}{*}{Mask efficiency $\uparrow$} & \multicolumn{3}{c}{DressCode} & \multirow{2}{*}{VITON-HD}  \\ \cmidrule(lr){2-4}
                    & dresses & upper body & lower body &          \\ \midrule
CatVTON             & 0.6876  & 0.8379     & 0.8179     & 0.6877   \\ 
IDM-VTON            & 0.7334  & 0.8196     & 0.8238     & 0.6889   \\ 
InstructVTON        & \textbf{0.8269} & \textbf{0.8924} & \textbf{0.8653} & \textbf{0.7808} \\ 
\bottomrule
\end{tabular}
}
\caption{Mask efficiency on each category of DressCode dataset. Mask efficiency is measured by the ratio of the human model image preserved by the mask. Given the same quality of try-on result, lower ratio of masked area is better since more pixels will be preserved.}
\label{tab:mask-efficiency}
\end{table}

\begin{table}[h!]
\centering
\resizebox{\linewidth}{!}{
\begin{tabular}{ccccc}
\toprule
\multirow{2}{*}{SSIM $\uparrow$} & \multicolumn{3}{c}{DressCode} & \multirow{2}{*}{VITON-HD} \\ \cmidrule(lr){2-4}
                    & dresses & upper body & lower body &          \\ \midrule
CatVTON             & 0.8878  & 0.9368     & \textbf{0.9294}     & \textbf{0.9186}   \\ 
IDM-VTON            & 0.9062  & 0.9294     & 0.9174     & 0.9096   \\ 
InstructVTON        & \textbf{0.9078} & \textbf{0.9370} & 0.9213     & 0.8887   \\ 
\bottomrule
\end{tabular}
}
\caption{Try-on results measured by SSIM. Higher score indicates better quality.}
\label{tab:ssim}
\end{table}

\begin{table}[h!]
\centering
\resizebox{\linewidth}{!}{
\begin{tabular}{ccccc}
\toprule
\multirow{2}{*}{LPIPS $\downarrow$} & \multicolumn{3}{c}{DressCode} & \multirow{2}{*}{VITON-HD} \\ \cmidrule(lr){2-4}
                    & dresses & upper body & lower body &          \\ \midrule
CatVTON             & 0.1269  & 0.0597     & 0.0788     & 0.0918   \\ 
IDM-VTON            & 0.0803  & 0.0528     & 0.0804     & \textbf{0.0706}   \\ 
InstructVTON        & \textbf{0.0689} & \textbf{0.0478} & \textbf{0.0678} & 0.0874   \\ 
\bottomrule
\end{tabular}
}
\caption{Try-on results on VITON-HD and each category of DressCode dataset, measured by LPIPS. Lower score indicates better quality.}
\label{tab:lpips}
\end{table}

\subsection{Qualitative results of multi-garment try-on}

\begin{figure*}[hbt!]
    \centering
    \includegraphics[width=0.9\textwidth]{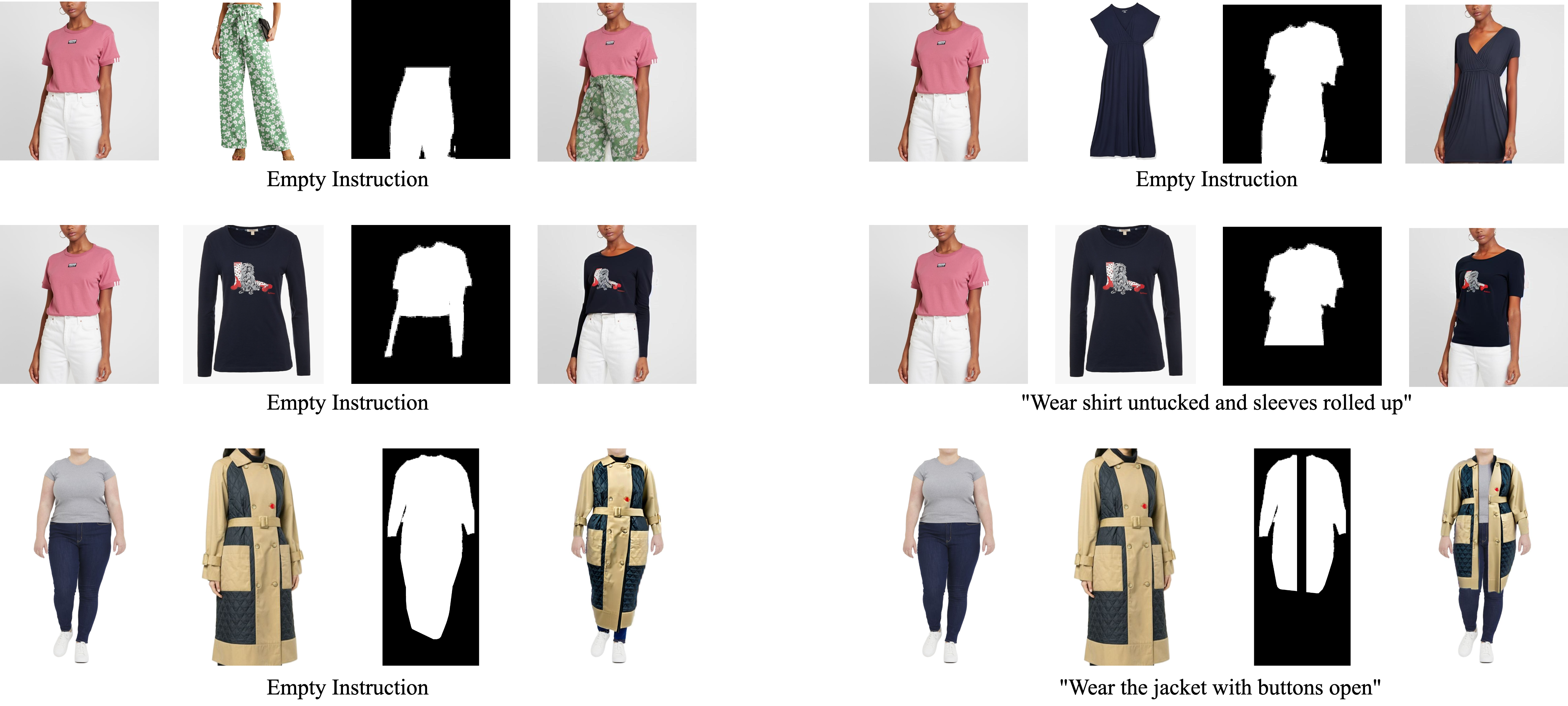}
    \caption{AutoMask and try-on result generated based on single-pair of human model image and target garment image with instruction. InstructVTON infers the most intuitive layout when instruction is empty.}
    \label{fig:single-pair-ins-noins}
\end{figure*}

\begin{figure}[hbt!]
    \centering
    \includegraphics[width=0.9\columnwidth]{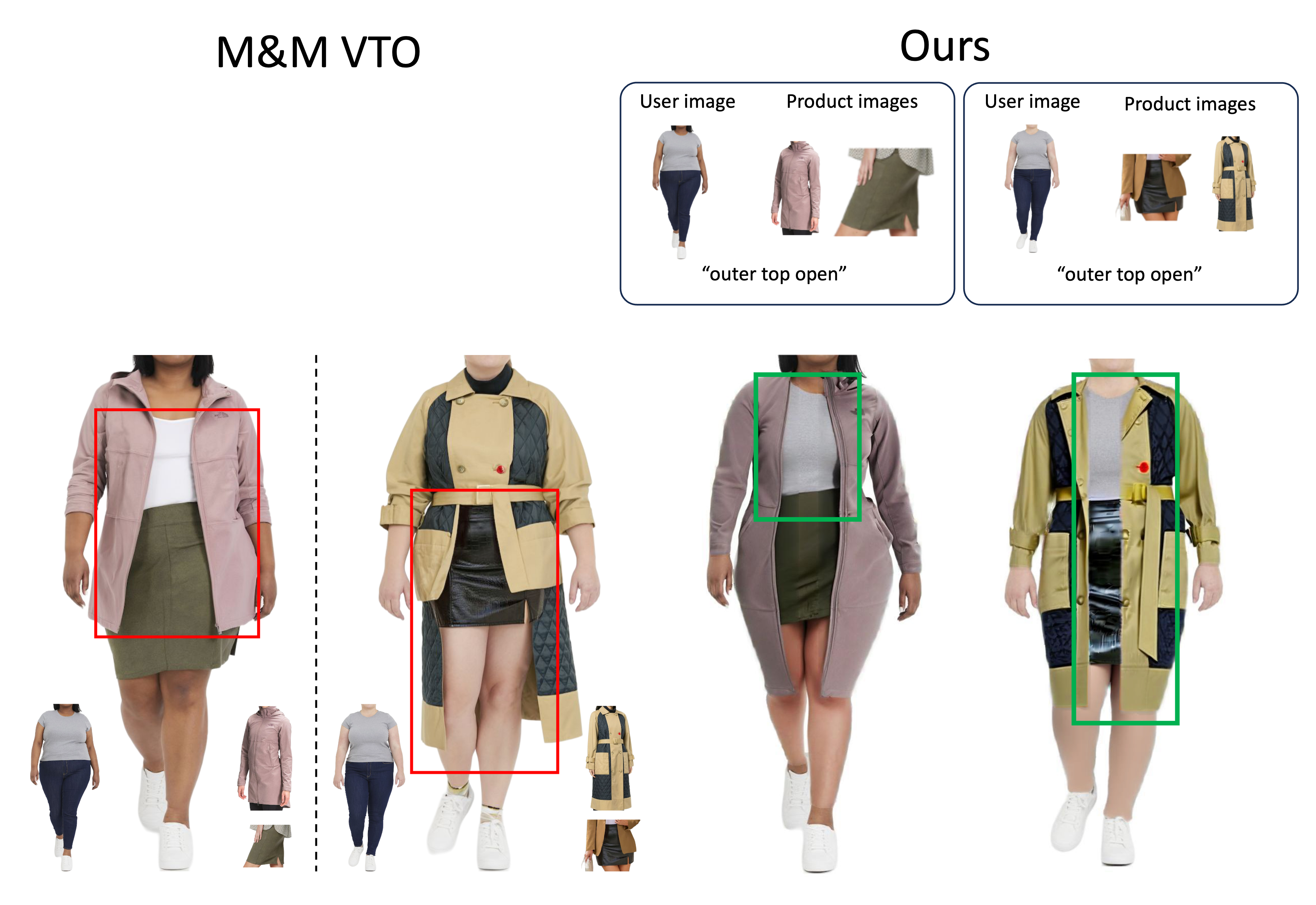}
    \caption{Comparison of InstructVTON on failure cases reported from M\&M VTO~\cite{10655292} (Figure to the left reproduced from~\cite{10655292}). On the left M\&M VTO generated white shirt under the open coat and struggled with rare garment combination.}
    \label{fig:direct-compare}
\end{figure}



Figure \ref{fig:single-pair-ins-noins} shows InstructVTON-generated masks and final try-on results of pairs of human model image and target garment image following different instructions. InstructVTON autonomously infers the optimal masking area in the human model image based on the type of target garment and style instruction. Notably, when the target garment is an overcoat, InstructVTON generated a mask that covers the area in between the legs to produce natural looking try-on result. For “wear the jacket with buttons open” style instruction, it removes a stripe from the center of the masking area to lead the inpainting VTO model to generate open-chest style try-on result while fully preserves the original garment underneath. We also qualitatively test InstructVTON against failure cases reported in~\cite{10655292}, see Figure \ref{fig:direct-compare}. ~\cite{10655292} uses a personalization-based approach which is different from the InstructVTON approach, as the former personalizes the model to the new human figure, while InstructVTON targets zero-shot application on any new source figure.

\section{Limitations and future work}

While our approach achieves complex VTO scenarios, it still has some limitations. 

First, The main limitation of our approach is latency. For a scenario with 3 target garments, the agent takes around 1 minute using state-of-the-art tools (segmentation, inpainting VTO, VLM), due to multiple calls to these intermediate models. Therefore, our approach is only suitable for offline use-cases where real-time interactivity is not a requirement. One way we plan to overcome this limitation is to distill the InstructVTON agent end-to-end into a single end-to-end model, using InstructVTON as teacher and the end-to-end model as a student. We are actively working on this approach.

Second, Our implementation of AutoMasking uses rough segmentation of the human body parts up to a limit granularity, which limits the accuracy and flexibility of the styling control in certain situations. For example, the final mask either includes or excludes the lower arms from the body parts segmentation to achieve sleeves rolled down or rolled up style, which may yield unsatisfying results when the user provides specific style instruction such as “rolling up the sleeves to 3 quarts length”. Higher granularity in body parts segmentations and AutoMask post-processing will enable more flexibility in styling control. 

Finally, both the agents in InstructVTON are open-loop planners that commit to a series of actions and execute them sequentially. An error in the earlier step can propagate through the entire plan and degrade the final results. In the future, we will explore modeling the InstructVTON agents with Markov decision process to mitigate error propagation and enable InstructVTON to handle more complex VTO use cases with uncommon garments and specific style instructions. We will also explore optimizing the InstructVTON agents with reinforcement learning.

{
    \small
    \bibliographystyle{unsrt}
    \bibliography{main}
}


\end{document}